# How Many Pages? Paper Length Prediction from the Metadata

Erion Çano
Institute of Formal and Applied Linguistics
Charles University, Prague, Czech Republic
+420 951 554 279
cano@ufal.mff.cuni.cz

Ondřej Bojar
Institute of Formal and Applied Linguistics
Charles University, Prague, Czech Republic
+420 951 554 276
bojar@ufal.mff.cuni.cz

## ABSTRACT
Being able to predict the length of a scientific paper may be helpful in numerous situations. This work defines the paper length prediction task as a regression problem and reports several experimental results using popular machine learning models. We also create a huge dataset of publication metadata and the respective lengths in number of pages. The dataset will be freely available and is intended to foster research in this domain. As future work, we would like to explore more advanced regressors based on neural networks and big pretrained language models.

## CCS Concepts
• **Information systems→Information extraction**
• **AppliedComputing→Document capture.**

## Keywords
page length prediction; corpus creation; research articles

## 1. INTRODUCTION
Many research papers from various disciplines are regularly published in online libraries. For example, the number of monthly submissions on Arxiv is currently higher than 16 thousand and is rapidly growing (June 2020 statistics from Arxiv website). One important aspect of a publication is its citation count dynamics in time which is being predicted using various techniques [1, 2]. Another important aspect is the relation between several attributes with each other and especially the way they statistically combine with stylistic metrics forming different writing styles [3]. There could be scenarios in which predicting the length of research papers based on their other attributes could be very helpful. Despite depending also on the layout, the length of a document in number of pages should correlate with other publication metadata and stylistic metrics as well. Understanding these latent relations could be useful for meta-research and important in vibrant applications such as plagiarism detection [4-6].

In this work, we focus on the length of the publications and propose a novel task as a regression problem: paper length prediction based on the metadata. We explored several online libraries and observed that many paper attributes are not always available. They still provide publication details such as *title*, *authors*, *abstract*, but *length* can be missing or hard to retrieve. To foster research in this direction, we crawled a big network of publication metadata [7] and created OAGL, a large dataset of paper attributes that we are freely releasing online.[1] It comprises about 17.5 million data samples with paper attributes and the corresponding length in number of pages, all stored as JSON lines. We also experimented with popular regression models on a small subset of OAGL to provide some initial baselines for the community. From our observations, basic regression models do not work well. However, ensemble models produce good results when their parameters are optimized. They also work better if trained with more features. Contrary, simple NN (Neural Network) models with static word embeddings are not very accurate. We believe that NN models based on big language models like BERT or GPT-2 [8, 9] that represent both words and contexts of text features (e.g., paper abstract) may provide better results.

## 2. OAGL DATASET CREATION
Creating and using datasets of scientific articles has become common recently [10-13]. There are several initiatives that crawl websites for integrating research resources in big and unified data networks. ArnetMiner [14] is one of such attempts that links together research data in a common network. One of its byproducts is the OAG (Open Academic Graph) data collection of scientific publications [7]. It is organized as a set of records containing article metadata like *title*, *authors*, *abstract*, *keywords*, *page length*, *publication year*, *isbn*, *issn*, *venue* and more. To produce an abundant collection of publication metadata and the respective page lengths, we used the OAG bundle. We decided to retrieve records with at least five categories which should be the most important: *title*, *keywords*, *abstract*, *publication year* and *page length*. Most of the obtained records do still contain other types of data like *number of citations*, *isbn*, *venue*, *volume*, etc.

Various publication records had very long or very short text attributes. For this reason, we ignored every record with a title not within 3 - 50 tokens, abstract not in the range of 40 - 400 tokens, keywords not within 2 - 20, and page length not in the range 2 - 50. Finally, we removed the duplicate entries and reached a total size of about 17.5 million records (precisely 17528680). Table 1 shows some statistics of the whole OAGL and the train, validation, and test splits (3500, 500, and 1000 samples each) we used for our experiments.[2] The titles and abstracts are on average 11.96 and 144.86 tokens long, with standard deviations 4.49 and 74.98 respectively. The number of keywords in each paper is also highly variable with a mean of 6.74 and a deviation of 5.49. The average paper length is 6.65 pages. We also noticed that about 90 % of the papers were published between 2000 and 2010. A data sample example from OAGL is illustrated in Table 2.

---

[1] OAGL is available at: http://hdl.handle.net/11234/1-3257
[2] Values of * attributes may vary based on the text preprocessing.

**Table 1. Statistics of the complete OAGL dataset and our experimentation splits**

| Attribute | Total | | Train | | Val | | Test | |
|---|---|---|---|---|---|---|---|---|
| | Mean | Std | Mean | Std | Mean | Std | Mean | Std |
| Title tokens* | 11.96 | 4.49 | 13.37 | 4.77 | 13.27 | 4.84 | 12.97 | 4.67 |
| Abstract tokens* | 144.86 | 74.98 | 159.01 | 65.09 | 155.35 | 61.06 | 154.53 | 59.02 |
| Keywords | 6.74 | 5.49 | 5.73 | 3.3 | 5.59 | 2.85 | 5.49 | 2.31 |
| Page length | 6.65 | 4.87 | 6.95 | 5.27 | 7.16 | 4.46 | 7.2 | 5.39 |

**Table 2. A data sample example from OAGL dataset**

"title": "Efficiency of wipe sampling on hard surfaces for pesticides and PCB residues in dust.", "abstract": "Pesticides and polychlorinated biphenyls (PCBs) are commonly found in house dust and have been described as a valuable matrix to assess indoor pesticide and PCB contamination. The aim of this study was to assess the efficiency and precision of cellulose wipe for collecting 48 pesticides, eight PCBs, and one synergist at environmental concentrations. First, the efficiency and repeatability of wipe collection were determined for pesticide and PCB residues that were directly spiked onto three types of household floors (tile, laminate, and hardwood). Second, synthetic dust was used to assess the capacity of the wipe to collect dust. Third, we assessed the efficiency and repeatability of wipe collection of pesticides and PCB residues that was spiked onto synthetic dust and then applied to tile. In the first experiment, the overall collection efficiency was highest on tile (38%) and laminate (40%) compared to hardwood (34%), p < 0.001. The second experiment confirmed that cellulose wipes can efficiently collect dust (82% collection efficiency). The third experiment showed that the overall collection efficiency was higher in the presence of dust (72% vs. 38% without dust, p < 0.001). Furthermore, the mean repeatability also improved when compounds were spiked onto dust (< 30% for the majority of compounds). To our knowledge, this is the first study to assess the efficiency of wipes as a sampling method using a large number of compounds at environmental concentrations and synthetic dust. Cellulose wipes appear to be efficient to sample the pesticides and PCBs that adsorb onto dust on smooth and hard surfaces.", "keywords": ["collection efficiency", "dust", "pesticides", "polychlorinated biphenyls", "wipes"], "year": 2015, "venue": "The Science of the total environment", "n citation": 16, "issn": "1879-1026", "volume": 505, "plength": 10

**Table 3. Different vectorizer scores with basic regression models**

| Vectorizer | Linear Regr | | | MLP Regr | | | SV Regr | | |
|---|---|---|---|---|---|---|---|---|---|
| | MSE | MAE | R2 | MSE | MAE | R2 | MSE | MAE | R2 |
| Tfidf | 30.01 | 3.9 | -0.03 | 28.71 | 3.81 | 0.01 | 28.03 | 3.29 | 0.04 |
| Hash | 32.37 | 4.15 | -0.11 | 29.24 | 3.87 | -0.06 | 26.84 | 3.19 | 0.08 |
| Count | 35.35 | 4.39 | -0.21 | 30.9 | 3.89 | -0.06 | 26.63 | 3.19 | 0.08 |
| Union | 35.21 | 4.38 | -0.21 | 29.53 | 3.82 | -0.02 | 26.63 | 3.12 | 0.08 |

# 3. OBSERVATIONS ON BASIC FEATURES

We ran several experiments with various regression models on a small subset of OAGL. At the beginning of each trial, we performed a few more processing steps, lowercasing the text fields and clearing the messy symbols in each sample. Furthermore, we used Stanford CoreNLP [15] to tokenize the titles and the abstracts. Our goal was to observe the role of different feature packs in the success of the length prediction task. The most important document attributes are the *title*, the *abstract* and the *keywords*. They are highly related with paper topics and should incorporate latent correlations with the page length. A primitive way to combine those three strings together is by simply concatenating them. We used different vector space models [16-18] for representing the common string and regression models for predicting the paper length.

In this first set of experiments, we vectorized the joint string of each paper record with *tfidf*, *count*, *hash*, and a *union* of the three of them. We also explored three machine learning models: an LR (Linear Regression), an SVR (Support Vector Regression) that uses the concept of support vectors [19] and an MLP (Multi-Layer Perceptron) for regression [20, 21] with their default parameters.[3] The respective MSE (Mean Squared Error), MAE (Mean Absolute Error) and R2 (R squared) scores are reported in Table 3. As we can see, the SVR performs better than the two other models. Regarding the vectorizers, *tfidf* performs best when combined with the LR and the MLP regressors. In the case of SVR, the *count* vectorizer leads. The union of the three does not seem to improve the feature extraction process. It is still worth to note that these observations are raw since no parameter optimization was performed, neither on the vectorizers nor on the regression models.

We ran a second set of experiments using two NNs on the same feature combination as above. The simplest model we tried is composed of an embedding layer for the text vectorization and a dense layer of 100 neurons followed by the output layer. We used static word embeddings of 300 dimensions from three sources: the

[3] https://scikit-learn.org/stable/supervised_learning.html

6 billion tokens collection of Common Crawl[4] trained with Glove [22], the 840 billion tokens collection of Common Crawl trained with Glove, and the 100 billion tokens collection of Google News[5] trained with Word2vec [23, 24]. The embedding layer is not trainable (we actually noticed that tuning the embeddings on our data negatively impacts performance) and serves only to create the vector space representation of the words. The maximal length of each word sequence was set to 400. As training optimizer we used Adam with its default parameters [25]. The training continued for 5 epochs with a batch size of 32.

The other NN structure is the NgramCNN architecture designed and used for sentiment analysis [26]. It is composed of an embedding layer for the word representation and several 1-dimensional convolution layers (feature extraction branches) of increasing filter sizes that extract unigrams, bigrams, trigrams or even longer word patterns (the W hyperparameter). The convolution layers are followed by max-pooling (or global max-pooling) layers and are repeated several times (the L hyperparameter). The branches are finally concatenated and a dense layer is used for regression, as it is illustrated in Figure 5 on page 12 of [26]. In this work, we used a very simple variant, with three branches of convolutions and a single pooling iteration (W =3 and L = 1). The embedding layer, and the training parameters were kept at the same values as in the other NN model. MSE, MAE and R2 scores for this second set of experiments are shown in Table 4. In general, we see that the scores are somehow better than those of Table 3. From the results, we notice that Glove embeddings perform better than word2vec ones. Regarding the two models, NgramCNN outruns the one-layer NN in all the three metrics.

**Table 4. NN and NgramCNN scores on static embeddings**

| Embeddings | OneLayerNN | | | NgramCNN | | |
|---|---|---|---|---|---|---|
| | MSE | MAE | R2 | MSE | MAE | R2 |
| CC6B-Glove | 25.55 | 3.61 | 0.12 | 24.68 | 3.23 | 0.15 |
| CC840B-Glove | 25.51 | 3.29 | 0.12 | 24.56 | 3.17 | 0.15 |
| Google-W2V | 26.1 | 3.3 | 0.1 | 25.06 | 3.2 | 0.14 |

# 4. ANALYZING MORE FEATURES

The scores reported in Tables 3 and 4 indicate that concatenating the *title*, the *abstract* and the *keywords* in a common string and vectorizing them together is not a good practice. Adding other paper metadata could also improve the regression results. For this reasons, we decided to run a third set of experiments adding publication *venue*, *year* and *citations* as extra features. The *venue* is a string indicating the conference or journal where the paper was published. The publication year and the number of citations are integers. Furthermore, we decided to vectorize the *title*, *abstract*, *keywords*, and *venue* independently using *tfidf* (the best vectorizer from the first set of experiments) and stacking them as columns in the feature matrix.

Once again, we used the LR, the SVR, and the MLP regressor but now we tried three ensemble models as well. An RF (Random Forest) is an example of a bagging ensemble method that aims to increase the strength and accuracy of learning algorithms [27, 28]. It runs in parallel and works well with different types of features. Contrary, boosting methods represent sequential ensembles that try to turn weak models into stronger ones by correcting the erroneous classifications of each iteration [29-31]. One of the most popular implementations is GB (Gradient Boosting) algorithm that is based on decision trees. XGBoost (Extreme Gradient Boosting) is a fast implementation that reduces the search space of possible feature splits [32]. The three of these ensemble methods work well on both classification and regression tasks. We examined the new feature pack of our OAGL subset using *tfidf* vectorizer and these six regression algorithms, trying to optimize their most important parameters. The results of the default models and of the optimized ones are presented in Table 5. Comparing the new scores of the LR, SVR, MLP models against the ones of Table 3, we notice considerable improvements. The LR and the MLP perform significantly better with new feature pack and are further improved by the parameter optimization process. The default SVR scores are slightly worse, but the optimized scores are significantly better, with R2 jumping up from -0.05 to 0.19.

**Table 5. Optimized model scores**

| Model | Default Params | | | GS Params | | |
|---|---|---|---|---|---|---|
| | MSE | MAE | R2 | MSE | MAE | R2 |
| LR | 23.89 | 3.45 | 0.18 | 22.54 | 3.3 | 0.22 |
| SVR | 30.43 | 3.51 | -0.05 | 23.58 | 3.14 | 0.19 |
| MLP | 24.19 | 3.39 | 0.17 | 22.72 | 3.26 | 0.22 |
| RF | 25.05 | 3.27 | 0.14 | 23.5 | 3.06 | 0.19 |
| GB | 22.44 | 3.14 | 0.23 | 21.6 | 3.04 | 0.26 |
| XGB | 22.36 | 3.12 | 0.23 | 21.16 | 3.05 | 0.27 |

The ensemble learners perform better, even with their default parameters. The RF is the weakest of the three, reaching an MSE of 23.5, an MAE of 3.06, and an R2 of 0.19 when optimized. GB and XGB perform similarly and reach optimized 21.6 and 21.16 MSE scores respectively. Moreover, XGB reached a 0.27 R2 score which is the highest we got in all the experiments. It is worth noting that XGB was not only the most accurate, but also the fastest ensemble learner. Furthermore, the parameter sets we searched were not exhaustive and further improvements could be achieved. Unfortunately, there are no literature baselines we could compare our results with. The optimal parameters we found for each model are presented in Table 6. Furthermore, we provide the source code to reproduce the experiments online.[6] We tried to further improve the results by adding some more statistical features like number of words in the title, number of words in the abstract, number of keywords, and number of capitalized words. There was no significant difference in the results, though. A final fact we observed was the insignificant role of certain numeric scalers (we tried MinMaxScaler and MaxAbsScaler) on *year* and *citations* features.

[4]https://nlp.stanford.edu/projects/glove/

[5]https://code.google.com/archive/p/word2vec

[6]https://github.com/erionc/paper-length

**Table 6. Top gridsearch parameters of the vectorizer and regressor in each model**

| Model | | Optimal Parameter Values |
|---|---|---|
| LR | vec | ngram range: (1,3), norm: l2, smooth idf: True, stop words: None, sublineartf: True |
| | reg | copy X: True, fit intercept: True, normalize: False |
| SVR | vec | ngram range: (1,3), norm: None, smooth idf: True, stop words: None, sublineartf: True |
| | reg | C: 10, gamma: auto, kernel: poly, shrinking: True |
| MLP | vec | ngram range: (1,2), norm: l2, smooth idf: True, stop words: None, sublineartf: True |
| | reg | hidden layer sizes: (100, ), alpha: 0.00005, solver: adam |
| RF | vec | ngram range: (1,3), norm: None, smooth idf: True, stop words: None, sublineartf: False |
| | reg | n estimators: 60, max features: auto, bootstrap: True, oob score: True |
| GB | vec | ngram range: (1,3), norm: None, smooth idf: True, stop words: None, sublineartf: True |
| | reg | n estimators: 100, max features: auto, max depth: 6 |
| XGB | vec | ngram range: (1,3), norm: None, smooth idf: True, stop words: None, sublineartf: False |
| | reg | n estimators: 70, eta: 0.008, gamma: 0.15, max depth: 6 |

# 5. CONCLUSIONS AND FUTURE WORK

In this paper, we proposed a novel task: predicting paper length using various publication details as features. We also created a large dataset of publication metadata that will be freely available. It is intended to encourage experimentation with various types of predictive models on this research direction. From our initial experiments, we noticed that basic regression models are not very accurate, leading to error rates that are relatively high.

Optimized ensemble models work better and may produce satisfying results with better feature processing and combinations. As future work, we would like to try neural network structures based on pretrained language models that are becoming very popular in language-related tasks. Given the large size of the data we dispose, we also want to examine the task from the data efficiency viewpoint [33], checking the scalability of the prediction scores when more training sample are used. A deeper understanding of the hidden relations between document length, publication attributes and other writing metrics could be invaluable for many applications and tasks.

# 6. ACKNOWLEDGMENTS

This research work was supported by the project no. 19-26934X(NEUREM3) of the Czech Science Foundation and ELITR (H2020-ICT-2018-2-825460) of the EU.